\documentclass[letterpaper, 10 pt, conference]{ieeeconf}  

\usepackage{graphics} 
\usepackage{amsmath,graphicx}
\usepackage{amssymb}
\usepackage{array}
\usepackage{graphicx}
\usepackage{booktabs}
\usepackage{url}
\usepackage{color}
\usepackage{bm}

\title{\LARGE \bf
What evidence does deep learning model use to classify Skin Lesions?
}

\author{Xiaoxiao Li$^{1, \dagger}$, Junyan Wu$^{2, \dagger,\star}$, Eric Z. Chen$^{3, \dagger}$ and Hongda Jiang$^4$}  

\begin{document}

\maketitle
\footnotetext[1]{Xiaoxiao Li,
        Yale University, New Haven, CT, USA.} 
\footnotetext[2]{Junyan Wu,
        Cleerly Inc, New York City, New York, USA.}
\footnotetext[3]{Eric Z. Chen,
        Dana-Farber Cancer Institute, Boston, MA, USA.}
\footnotetext[4]{Hongda Jiang,
        East China University of Science and Technology, Shanghai, China.}%
\footnotetext[5]{$^{\star}$ To whom correspondence should be addressed}
\footnotetext[6]{$^{\dagger}$ These authors contributed equally.}

\thispagestyle{empty}
\pagestyle{empty}

\begin{abstract}
Melanoma is a type of skin cancer with the most rapidly increasing incidence. Early detection of melanoma using dermoscopy images significantly increases patients' survival rate. However, accurately classifying skin lesions by eye, especially in the early stage of melanoma, is extremely challenging for the dermatologists. Hence, discovery of reliable biomarkers will be meaningful for melanoma diagnosis. Recent years, the value of deep learning empowered computer assisted diagnose has been shown in biomedical imaging based decision making.  However,  much research focuses on improving the disease detection accuracy but not exploring the evidence of pathology.  In this paper, we propose a method to interpret the deep learning classification findings. Firstly, we propose an accurate neural network architecture to classify skin lesions. Secondly, we utilize a prediction difference analysis method that examines each patch on the image through patch-wised corrupting to detect the biomarkers. Lastly, we validate that our biomarker findings are corresponding to the patterns in the literature. The findings can be significant and useful to guide clinical diagnosis.
\end{abstract}
%
\section{Introduction}
\label{sec:intro}
Skin cancer is a severe public health problem in the United States, with over 5,000,000 newly diagnosed cases every year. Melanoma, as the severest form of skin cancer, is responsible for $75\%$ of deaths associated with skin cancer \cite{jerant2000early}. In 2015, the global incidence of melanoma was estimated to be over 350,000 cases, with almost 60,000 deaths. Fortunately, if detected early, melanoma survival exceeds $95\%$ \cite{ISIC2018}. 

Dermoscopy is one of the most widely used skin imaging techniques to distinguish the lesion spots on skin \cite{binder1995epiluminescence}.
It has been developed to improve the diagnostic performance of melanoma. The manual inspection from dermoscopy images made by dermatologists is usually time-consuming, error-prone and subjective (even well-trained dermatologists may produce widely varying diagnostic results)\cite{binder1995epiluminescence}. 
Nevertheless, the automatic recognition of melanoma using dermoscopy images is still a challenging task due to the following reasons: the low contrast between skin lesions and normal skin regions makes it difficult to accurately segment lesion areas; the melanoma and non-melanoma lesions may have a high degree of visual similarity;  the variation of skin conditions such as skin color, natural hairs or veins, among patients produce the different appearance of melanoma, in terms of color and texture, etc. Some investigations attempted to apply low-level hand-crafted features to distinguish melanomas from non-melanoma skin lesions \cite{yang2018clinical}. Recent works employing Convolutional Neural Networks (CNNs) have shown its improved discrimination performance in melanoma classification aiming at taking advantage of their discrimination capability to achieve performance gains \cite{yu2017automated}. Although these studies focused on improving computer assisted diagnostic accuracy, the diagnosis itself is hard even for experienced clinical practitioners based on dermoscopy images. The computer intervention not only assist decision making, but also can benefit clinical research to identify the biomarkers which contribute to diagnosing. Despite promising results, the clinicians typically want to know if the model is trustable and how to interpret the results. Biomarker interpretation from deep learning models for clinical use has been explored in identifying brain disease \cite{li2018brain,zintgraf2017visualizing}. However, to the best of our knowledge, what kind of evidence deep learning models use for classifying skin lesions has not been explored.
Experienced dermatologists diagnose skin diseases based on comprehensive medical criteria which have been
verified to be useful, e.g., the ABCD rule \cite{abbasi2004early} and the 7-point checklist \cite{walter2013using}, etc.  There is still much
room to improve the understanding of melanoma recognition by reliable CNNs classifier. In this paper, we proposed a novel method based on deep convolutional neural networks and low-level image feature descriptors, which imitate
clinical criteria representations, to solve skin lesion analysis towards melanoma
detection problem. We aim to inspect whether the deep learning models and the dermatologists use similar criteria. 
There are three main approaches for interpreting the important features detected by DNNs. One approach is using gradient ascent methods to generate an image that best represents the class \cite{yosinski2015understanding}. However, this method cannot handle nonlinear DNNs well.  The second one uses the intermediate outputs of the network to visualize the feature patterns \cite{zhou2016learning}, but ends up with blurred heatmaps.  The third approach is to visualize how the network responds to a specific corrupted input image in order to explain a particular classification made by the network \cite{zintgraf2017visualizing}, which can more precisely locate the key features and do not need to retrain the network. 


Here, we focus on interpreting the import features used in CNNs. Generally speaking, our interpretation method belongs to the third approach.
Therefore, we propose a pipeline to identify the evidence and biomarkers in the skin lesion dermoscopic images, which contributes to the deep learning classifier. To be specific, we first trained an accurate deep learning model to classify each dermoscopic image, with a predicted probability score to each class. Secondly, we analyzed the feature importance by corrupting with conditional sampling, then compare the prediction difference. 

\begin{figure*}[htpb]
	\centering
    \centerline{\includegraphics[width=16cm]{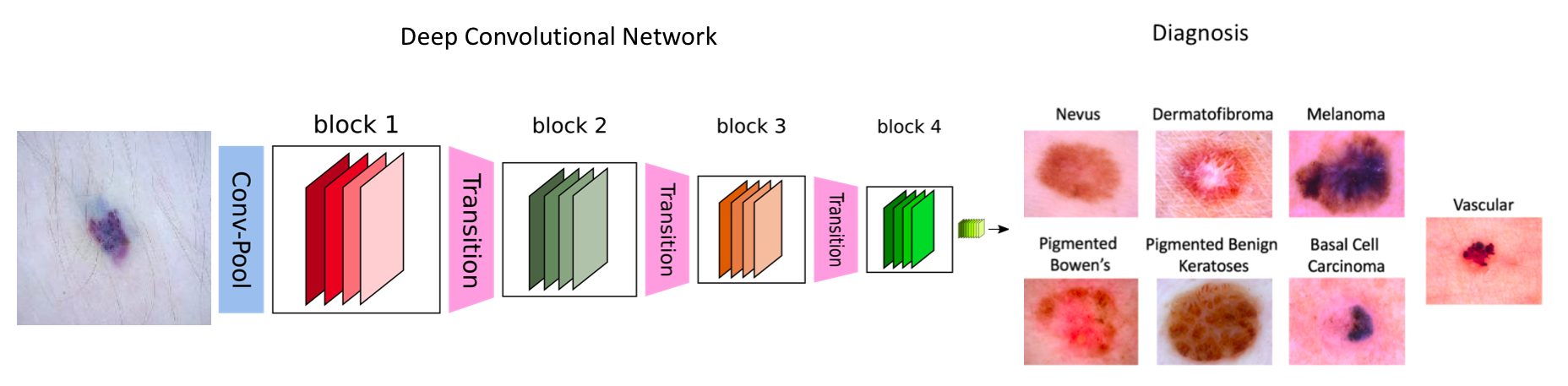}}
     \caption{The flowchart of deep learning model to classify skin lesions \cite{ISIC2018}}
    \label{pipeline}
\end{figure*}

\section{Method}
\label{sec:format}

\subsection{Deep Learning Image Classifier}
CNNs have led to breakthroughs in natural images classification and object recognition. Since CNNs have hierarchical feature learning
capability,  they have been widely used in natural images classification and object recognition, due to their hierarchical feature learning capability and discrimination performance. For instance, the CNN based methods outperform the traditional techniques significantly in the recent ImageNet challenges \cite{DengCHI14}. Skin lesions have large inter classes variations. Hence the key distinguishable features that differentiate the skin lesions are difficult to be fully captured by traditional computer vision approach. We modified and fine-tuned two successful image classification CNNs architectures: ResNet50 \cite{he2016deep} and VGG \cite{simonyan2014very} to  encode the image features. The fully
connected(FC) layers of both networks were modified by removing the last FC
layer and adding FC layer with 128 kernels and FC layer with 7 (number of
classes) for the $7$ class skin lesions classification task. We used LightGBM \cite{ke2017lightgbm} to combine the different CNN model features. The LightGBM is a boosting tree-based learning algorithm. It provides multiple hyper-parameters for achieving best performance.  

\subsection{Interpreting Deep Learning Features}
\label{diffmethod}
The interpretation method we adopted here is based on \cite{zintgraf2017visualizing}, which visualize how the CNNs respond to a specific corrupted input image in order to explain a particular classification made by the network. This method can localize the features contributing to the classifier. We use a heuristic method to estimate the feature (an image patch) importance by analyzing the probability of the correct class predicted by the corrupted image. Simply speaking, we corrupt the pixels in a sliding window which covers the region of interest (ROI), and then analyze the difference of prediction outcome. In this way, the relevance of features can be estimated by measuring how the prediction changes if the feature is corrupted. In the deep learning classifier case, the probability of the abnormal class $c$ given the original image $\bm{X}$ is estimated from the predictive score of the model $f$ : $ f(\bm{X}) = p(c|\bm{X})$. We denote the image corrupted at ROI $\bm{i}$ as $\bm{X_{\setminus i}}$. The prediction of the corrupted image is $p(c|\bm{X_{\setminus i}})$. To calculate $p(c|\bm{X_{\setminus i}})$, we marginalize out the corrupted ROI $\bm{i}$:
\begin{equation}
p(c|\bm{X_{\setminus i}}) = \mathbb{E}_{\bm{x_i}\sim p(\bm{x_i}|\bm{X_{\setminus i}})}p(c|\bm{X_{\setminus i}},\bm{x_i}),
\end{equation}
where $\bm{x_i}$ is a sample of feature $i$. Once the class probability $p(c|\bm{X_{\setminus i}})$ is estimated, we compare it with the original prediction $p(c|\bm{X})$ by calculating \textit{Weight of Evidence (WE)}:
\begin{equation}
    WE_i(c|\bm{X})= log_2(odds(c|\bm{X})) - log_2(odds(c|\bm{X_{\setminus i}}))
\end{equation}
where $odds(c|\bm{X}) = p(c|\bm{X})/(1-p(c|\bm{X}))$. Laplace correction ($p \longleftarrow (pN+1)/(N+K)$) is used to avoid zero probabilities, where $N$ is the number of training instances and $K$ is the number of classes. $WE$ can be positive and negative. The feature with positive value stands for the evidence for the classifier. The feature with negative value may be the common features shared in different classes, which confuses the classifier. 

Following \cite{zintgraf2017visualizing}, the corrupted $ROI_i$  ($win\_size \times win\_size$) will be replaced by conditionally sampled with $x_i$ from its surrounding neighbours. Therefore, we apply a larger patch $\mathbf{\hat{x}_i}$ ($size = win\_size + pad\_size$) including feature $x_i$, approximating the generative model as:
\begin{equation}
    p(x_i|\bm{X_{\setminus i}}) \approx p(x_i|\bm{\mathbf{\hat{x}_i}_{\setminus i}})
\end{equation}where $p(x_i|\bm{\mathbf{\hat{x}_i}_{\setminus i}})$ is a multi-Gaussian model of the whole patch (window + padding). In order to examine all the feature patterns in the image, the patch was applied as a sliding window to traverse the whole image with overlapping.

\begin{figure}[htpb]
	\centering
    \centerline{\includegraphics[width=9cm]{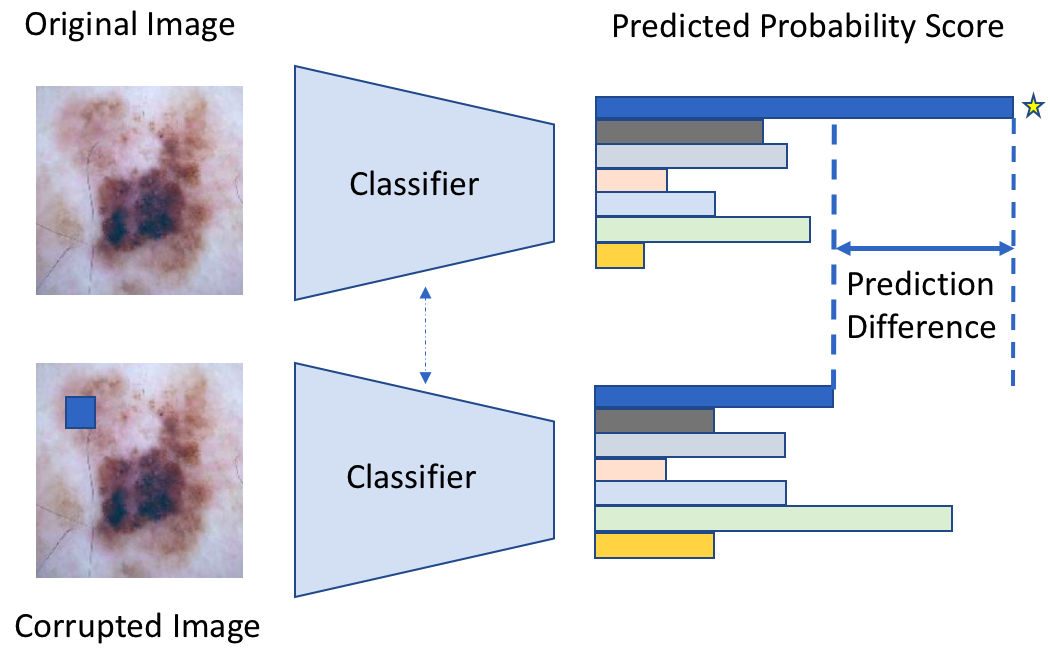}}
     \caption{Prediction difference analysis. Top: The prediction scores for each class given by the CNN classifier on the original image. The original image belongs to the class $c$ denoted with a star. Bottom: The prediction scores for each class given by the same CNN classifier on the same image where the blue region (ROI) was corrupted. The difference in the two prediction scores illustrate the importance of the blue region in the classification decision made by the CNN classifier.} 
     
    \label{fig_diff}
\end{figure}
\section{Experiments and Results}

\subsection{Dataset}
The dataset was extracted from the open challenge dataset of Skin Lesion Analysis Towards Melanoma Detection (ISIC 2018) \cite{tschandl2018ham10000,codella2018skin},which were collected from leading clinical centers internationally and acquired from a variety of devices within each center. Broad and international participation in image contribution is designed to ensure a representative clinically relevant sample. It consisted of 10015 images (327 actinic keratosis (AKIEC), 514 basal cell carcinoma (BCC), 115 dermatofibroma (DF), 1113 melanoma (MEL), 6705 nevus (NV), 1099 pigmented benign keratosis (BKL), 142 vascular lesions (VASC)). Each data is RGB color image, with size $450 \times 600$.

\subsection{Deep learning classification}
\label{sec:pagestyle}

We used 3 models: VGG16, ResNet50 and ensembled VGG16 + ResNet50 to classify the resized images. The loss function optimized to train the networks was categorical cross-entropy. 

We split $70\%$ data as training set, $10\%$ as the validation set, which was used to find the early stopping epoch and $20\%$ as testing set to evaluate our algorithms. The number of testing samples in each class were listed in Tabel \ref{all}.

As the number of images in each category varies widely and unbalanced, we augmented the images of different classes in the training set accordingly. The augmentation methods included randomly rotation up to $25^o$, left-right flipping, top-bottom flipping and zoom-in cropping with ratio 0.8. All the input images were re-sized to (224, 224) in our application.

The performance of each model is summarized in Table \ref{compare}. We observed that the ensemble model outperformed the single model. The feature interpretation analysis described in the next subsection was applied on the ensemble model, since the more accurate the classifiers were, the more reliable the interpreted results were. The classification results of VGG16 + ResNet50 ensemble model were summarized in Table \ref{all}.

\begin{table}

  \caption{Comparison of different model strategies}
  \label{sample-table}
  \centering
  \begin{tabular}{p{1.5cm}  p{1.5cm}<{\centering} p{2cm}<{\centering} p{2cm}<{\centering}} 
    \toprule
    Model     & VGG & ResNet & \textbf{VGG+ResNet}  \\
    \midrule
    Accuracy   & 0.79 & 0.82  & \textbf{0.85 } \\
    \bottomrule
  \end{tabular}
  \label{compare}
\end{table}

\begin{table}
  \caption{Performance summary of the ensemble model}
  \label{summary-training}
  \centering
  \begin{tabular}{p{1.5cm} p{1cm}<{\centering} p{1cm}<{\centering} p{1.5cm}<{\centering} p{1cm}<{\centering}}
  \toprule
  Categories & Precision & Recall & F1 score & Samples \\
  \midrule
  MEL   & 0.70  & 0.54 & 0.61 & 223\\
  NV    & 0.90  & 0.96 & 0.93 & 1341 \\
  BCC   & 0.78  & 0.78 & 0.78 & 103 \\
  AKIEC & 0.56  & 0.61 & 0.58 & 66 \\
  BKL   & 0.76  & 0.66 & 0.71 & 220 \\
  DF    & 0.83  & 0.65 & 0.73 & 23 \\
  VASC  & 0.84  & 0.72 & 0.78 &  29 \\
  \midrule
  Total & 0.84 & 0.85 & 0.84 & 2005\\
  \bottomrule
  \end{tabular}
  \label{all}
\end{table}
\subsection{Deep learning feature interpretation}
\label{sec:typestyle}
\begin{figure}[htpb]
	\centering
    \centerline{\includegraphics[width=10cm]{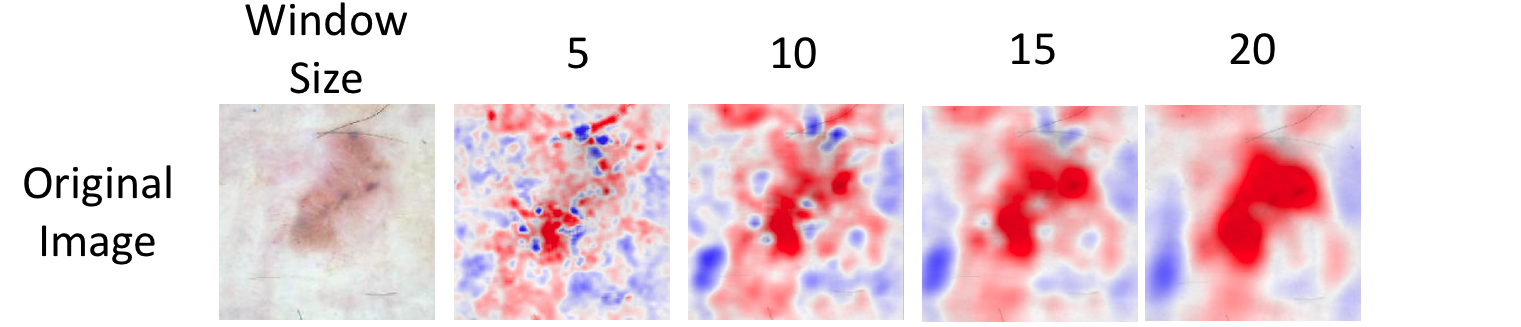}}
     \caption{Investigating different window sizes to interpret the skin lesion classification results. }
    \label{win}
\end{figure}

\begin{figure}[htpb]
	\centering
    \centerline{\includegraphics[width=9cm]{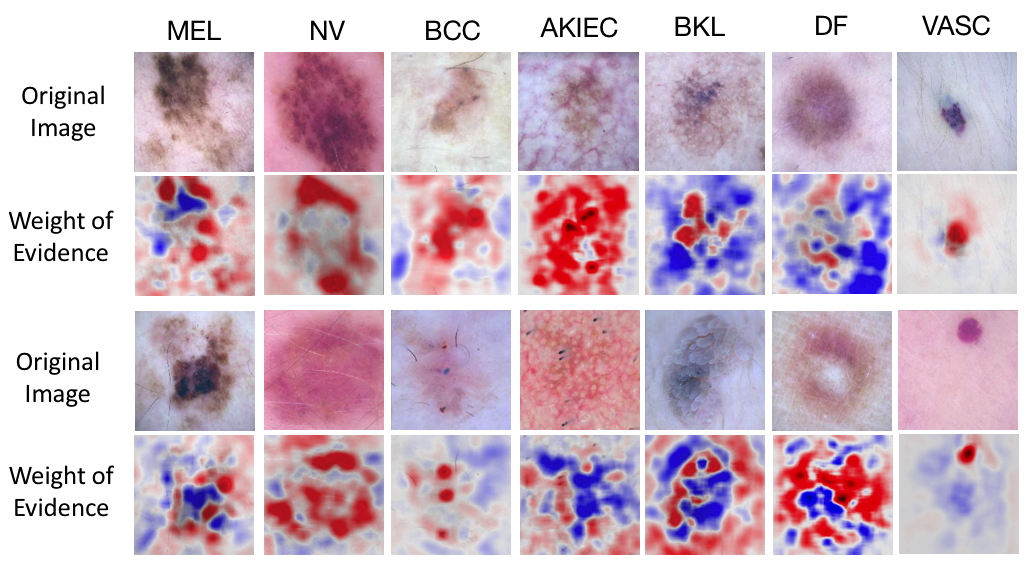}}
     \caption{The interpretation results ($win\_size = 15$) for classifying skin lesion. Each column stands for each class. There are seven classes in our classification task.  The original images are shown in the $1^{st}$ and $3^{rd}$ rows, where two instances of each class are given. Their weights of evidence maps are shown in the $2^{nd}$ and $4^{th}$ rows correspondingly. Red color highlights the evidence for the classifier and blue color highlights the evidence against the classifier.}
    \label{diff}
\end{figure}

In order to interpret the features the deep learning classifier used to classify skin lesions, we did the prediction difference analysis as described in section \ref{diffmethod}. We investigated the ROI with length $win\_size$ in a patch with length $win\_size + pad\_size$. This patch traversed around the whole image with overlapping. Notably, each pixel of the image was visited multiple times $T$, except the 4 pixels on the four corners of the image. The final \textit{weight of evidence} assigned to pixel $p$ is $WE_p(c|\bm{X}) = \sum_t^T WE_p^t(c|\bm{X})$. We set the size of padding to 2, which was used to find the surrounding pixels around the feature and generate the Gaussian parameters for conditional sampling. For visualization, we colored pixels with \textit{positive WE} in red and pixels with \textit{negative WE} in blue. We investigated the suitable $win\_size$ to capture the predictive feature for the deep learning classifier. We set $win\_size = 5, 10 ,15$ and $20$. The results are shown in Fig. \ref{win}.
We found that, with $win\_size = 10$ or $15$, most distinguishing features were captured and we could get interpretable results. Hence, we chose $win\_size = 15$ and randomly displayed the two instances of each class in Fig. \ref{diff}. 

\subsection{Discussion}
\label{sec:majhead}
From the two instances for each class given in Fig. \ref{diff}, we observed that the features contributing to the classifier (highlighted in red) followed the patterns below. Further, we validated the results and asked for opinions from 2 dermatologists in the different institutions. For MEL, the neural network marked dark, dense, variously sized, asymmetric distributed structures, as validated by dermatologists that Blue white veils, dark brown homogeneous areas that lack pigment network were picked which are the clinical evidence for MEL, suggested by the dermatologists. For NV, the pigment network and the globular structure are marked, corresponding to the clinical evidence \cite{NV} and validated by the dermatologists. In addition, the boundary of the lesion was marked, where the size of the lesion might be evidence for the classifier. As for BCC, small gathered spots were marked, which are blue gray globular  pointed out by the dermatologists. In AKIEC class, the dermatologists pointed out the annular–granular structures of the skin and hair follicle openings surrounded by a white halo were marked by red, which are clinical evidence for AKIEC class \cite{AKIEC}. Small nub-like extensions were highlighted in BKL\cite{BKL}. The dermatologists also comfirmed that DF was marked on the peripheral delicate pigment network \cite{DF}. For VASC, circumscribed and ovoid structures  \cite{VASC}, which were thought as lucunae of the blood vessels by the dermatologists were marked. Also, it seems like the classifier barely considered the surrounding skin information when recognizing VASC.  It is interesting to see that surrounding skins can be used as evidence to classify skin lesions in CNNs. The blue regions were either background or the common features shared by different classes that negatively impact the classifier.

\section{Conclusion}
\label{ssec:subhead}
In this paper, we proposed a pipeline to interpret the saliency features (biomarkers) detected by deep learning model to classify skin lesion dermoscopic images. 
Our ensemble CNNs classifier conducted on the open challenge dataset of Skin Lesion Analysis Towards Melanoma Detection (ISIC 2018) demonstrated the effectiveness of the proposed method, even with limited and unbalanced training data. From the interpreted weight of evidence maps, we found discernible features of each class. The patterns match the dermatologist criteria identifying potential for improving clinical skin lesions detection. Further work will improve the classifier's performance, including investigating integrating probability map generated
by segmentation in our networks, will try more interpretation methods (such as LIME \cite{ribeiro2016should}, DeepLift \cite{shrikumar2017learning}, and Integrated Gradients \cite{sundararajan2017axiomatic}) and will explore more applications in different disease .


\bibliographystyle{IEEEbib}
\bibliography{root}

\end{document}